# Wikipedia for Smart Machines

# and Double Deep Machine Learning

**Moshe BenBassat\* (**moshe.benbassat@plataine.com**)**

**Arison School of Business, Interdisciplinary Center (IDC), Herzliya, Israel**

If you wish to read more about my AI work and earlier publications, please visit www.moshebenbassat.com, contact me at: moshe.benbassat@plataine.com and follow me on LinkedIn: www.linkedin.com/in/moshe-benbassat-AI.

## Abstract

Very important breakthroughs in **data-centric** deep-learning algorithms led to impressive performance in 'transactional' point applications of Artificial Intelligence (AI) such as Face Recognition, or EKG classification. With all due appreciation, however, **'knowledge-blind' data-only** machine learning algorithms have severe limitations for non-transactional AI applications, such as medical diagnosis beyond the EKG results. Such applications require deeper and broader knowledge in their problem-solving capabilities, e.g. integrating anatomy and physiology knowledge with EKG results and other patient's findings. Following a review and illustrations of such limitations for several real-life AI applications, we point at ways to overcome them.

The proposed **Wikipedia for Smart Machines** initiative aims at building repositories of software structures that represent humanity's science & technology knowledge in various parts of life; knowledge that we all learn in schools, universities and during our professional life. Target readers for these repositories are smart machines; not human. AI software developers will have these Reusable Knowledge structures readily available, hence, the proposed name **ReKopedia**. Big Data is by now a mature technology, it is time to focus on 'Big Knowledge'. Some will be derived from data, some will be obtained from mankind's gigantic repository of knowledge.
**Wikipedia for smart machines** along with the new **Double Deep Learning** approach offer a paradigm for integrating data-centric deep learning algorithms with algorithms that leverage deep knowledge, e.g. evidential reasoning and causality reasoning. The resulting synergies establish broader and deeper foundations that will enable us to scale faster the AI field. For illustration, a project is described to produce ReKopedia knowledge modules for medical diagnosis of about 1,000 disorders.

We are now in the second AI 'spring' after a long AI 'winter'. To avoid sliding again into an AI winter, it is essential that we rebalance the roles of data and knowledge. **Data is important, but knowledge- deep, basic, and commonsense- is equally important.**

\*See personal note, acknowledgements and a short bio at end of article.

## 1. Introduction

In recent years, Deep Learning (DL) and other machine learning (ML) algorithms achieved very important breakthroughs and outstanding results [14], [9]; primarily in image, speech and natural language understanding; leading to impressive performance in 'transactional' tasks such as bank-check verification, detecting anger, alerts from a Face Recognition system, or EKG interpretation. DL algorithms use data to automatically train neural networks to make intelligent inferences for tasks covered by the training data. They are based on mathematical/statistical models which means that their behavior is predictable. Once trained to a certain level, they are likely to perform consistently- within their statistical error boundaries- for cases within the scope of their training data. Another key advantage of DL algorithms is the 'no human touch feature engineering' for pattern recognition tasks, meaning no need for lengthy research projects requiring domain experts, e.g. fingerprints experts, or EKG experts, to find/extract good differentiating features for classification decisions. Again, impressive achievements which are universal and domain-agnostic. During the early part of my AI career, I could have certainly benefited from DL in many pattern recognition projects, including: object recognition for a Ballistic Missile Defense system (during the Cold War era), ultrasound wave recognition for an autonomous machine digging coal on the moon, EEG/ERP waves, and handwritten character recognition, where we had to handcraft a good set of differentiating features. I view DL's remarkable progress and achievements as key development in the future of AI; and as a mathematician, it is music to my ears.

However, with all due appreciation, DL algorithms have their limitations, as discussed below, specifically with 'non-transactional' applications that require broader and deeper reasoning ('strong AI'); possibly involving multiple deep knowledge sources, e.g. optimizing manufacturing and service operations, medical diagnosis and equipment troubleshooting, or military situation assessment and mission planning. For example, sequential diagnosis and situation assessment require algorithms for **hypothesis generation**, **goal setting**, and human-oriented **information acquisition** in order to drive a focused cost-effective decision-making process [2], [8]. Knowledge-blind data-only DL algorithms are based on **reasoning by analogy and extrapolation,** and therefore are very limited in areas where deeper and broader reasoning are required. In fact, as Pearl points out in a recent article there are theoretical impediments to current machine learning techniques for **cause and effect reasoning** [16]. Both types of reasoning, by analogy and by causation, and others, are critically important for intelligent problem solving as elaborated later in the article.

Early successes with DL led to overstating its applicability, reaching claims such as 'with Deep Learning and sufficient amount of data you can solve all AI problems'. The discussion and examples bellow illustrate why being radically religious about such doctrine limits the progress of AI. Following an application-oriented non-technical discussion of DL's limitations, we proceed to introduce two concepts designed to enrich and amplify machine learning and AI in general by using the wealth of knowledge that mankind developed in many fields over thousands of years:

(1) The **Double Deep Learning** approach advocates integrating data-centric deep-learning techniques with machine-teaching of deep knowledge; like the difference between teaching physicians versus paramedics or teaching engineers versus technicians.
(2) The initiative for **Wikipedia for Smart Machines** aims at building repositories that contains software representation structures of humanity's science & technology knowledge in various parts of life. Target readers smart machines; not human. The goal is to develop methodologies, tools, and algorithms to convert humanity knowledge that we all learn in schools, universities and during our professional life into Reusable Knowledge structures that smart machines can use in their inference algorithms. AI software developers will have these knowledge structures readily available, hence, the proposed name is **ReKopedia**. Ideally, ReKopedia would be an open source shared knowledge repository similar to other open source software code repositories.

ReKopedia's knowledge foundations and Double Deep Learning algorithms establish together a paradigm for faster and stronger on-going machine learning based on the principle that 'the more you know the faster you learn'.

**Double Deep Learning** is likely to improve DL on both sides of the spectrum: increase quality and scope of **good** decisions- including for ambiguous cases in twilight zones- and, not less important, reduce the number of **glaring** mistakes, such as those we see by contemporary personal assistants like Siri, Cortana, and Alexa. (It is OK for an algorithm to make a mistake where a human professional may also err. It is totally unacceptable for an algorithm to make glaring mistakes that even a human beginner would not make. More research effort should be devoted to eliminating glaring mistakes by DL-based solutions as they raise very fundamental questions about the intelligence of the algorithm and risk its credibility). One way to start dealing with that, is to integrate at the output points of DL algorithms (or within their neural net?) **sanity checks (second opinion checks)** that are based **on external knowledge** to evaluate the sensibility of DL's output and, when needed, interfere with the ensuing actions before they are executed, e.g. mistakenly shutting down a nuclear reactor or a production line. Several examples are given in the next two sections.

The article is based on lessons I learned over decades of an AI-focused career. I avoid giving 'toy' examples. All examples below are based on- or inspired by- real-life non-transactional AI systems I deployed with my teams that benefit hundreds of millions of people around the globe (engineering, business, military, …). For example, the **service optimization software** we developed at ClickSoftware schedules daily about 750,000 Field Engineers (FE) for many service providers around the world, including some of the largest in utilities, telecommunications, office equipment and the like. Assuming that each engineer delivers on average 3 to 3.5 jobs per day, and works roughly 210 field days per year, this means that, over a year period, these products touch the life of about **500 Million people,** which are roughly 6% to 7% of the 7 Billion+ world population. Throughout the article I also use **medical** examples at a layman level. The main reason is that I found medicine to be the broadest common denominator for people and, in addition, I had the honor of spending more than 10 years of my career as one of the pioneers in developing AI solutions to support medical decision making, so I appreciate the complexity of AI challenges in this domain.

## 2. Deep Learning Limitations for AI Applications: A Non-Technical Overview

(a) **Transactional vs non-Transactional Tasks.** I used earlier the terms 'transactional', and 'non-transactional' tasks. Rather than going into formal definitions of these, and related terms such as 'strong AI' 'weak AI', let me use EKG interpretation versus medical diagnosis to clarify the difference.

**Example 1: EKG Classification versus Medical Diagnosis**. With thousands of EKG training data signals, a DL algorithm can do an excellent job classifying an EKG signal shape, e.g. producing as output: Class = "ST Elevation", see [18] for 'super-human' performance based on 64,000 EKG records. If the user now asks the DL algorithm to elaborate on the meaning of its output for patient diagnosis, prognosis and treatment, he/she is unlikely to receive a meaningful answer because the software has no clue what this funny signal shapes stand for. A physician, on the other hand, will explain that ST Elevation represents ventricular contraction and may indicate an artery clog that may damage the heart muscle (Myocardial Infarction). The difference between the DL algorithm and the physician is the level of understanding of EKG findings. The physician's answer is based on layers on top of layers of anatomy and physiology knowledge, including the electrical impulses and their relationship to EKG findings. For an AI-based medical diagnosis system, just EKG classification is a narrow point solution ('transactional') to a small part of the problem. To achieve diagnosis in the physician's sense of the word, the AI solution should go beyond signal analysis and connect a given EKG shape to the way the heart functions and fails and integrate these with other patient findings.
Can a DL algorithm learn from patient data ONLY the anatomy and physiology of the human heart? That is: learn the four chambers structure, the arteries, the valves, the conduction system and the pacemaker, the walls, …, the function of each module and the overall blood flow. **I doubt it simply because patient data does not contain the information to enable such learning**. (To appreciate the complexity, check https://www.youtube.com/watch?v=RYZ4daFwMa8 for an excellent heart simulation that also connects the human heart and EKG). Even with man-made fully documented equipment, e.g. printers or semiconductor equipment, the challenge for machine self-learning of equipment's structure, function, and process flow is enormous. For medical diagnosis the challenge is even greater because the 'engineer' of the human body did not leave us with any design documentation, ….
As a scientist, I am in favor of research to push further the spectrum of what data-only learning approaches can learn and understanding their boundaries. As a business executive and AI practitioner, I believe that producing today a working AI system for medical diagnosis, requires teaching computers explicitly the anatomy/physiology knowledge like we teach human medical students, rather than wait until, **and if**, a data-only DL algorithm will learn it from zero at a comparable level.

**Newspaper articles periodically report that DL has been successful in medical diagnosis, e.g. for cancer. Without taking anything away from the importance and value of these point solutions tools, a fairer description of the situation would be: DL algorithms support well narrow aspects of the medical diagnosis process by providing point solutions such as classifying an EKG signal, detecting tumors in an image, or searching for past patients that are most similar to a given patient.**

(b) **Prediction and Planning.** Planning for the future is a key element in business and life. Planning starts with predictions for potential future scenarios; including those that are the result of intervention actions or other 'surprise' reasons that may cause deviations from historic behavior, e.g. how will changes in import tax impact US recession? Prediction by a DL algorithm covers only future scenarios where **history is assumed to repeat itself** (statistically). Without the ability to analyze richer "what if" scenarios, AI tools to support planning would be very limited. By adding, for instance, the recently developed Transportability techniques of Causality theory [1], DL-based solutions can be expanded **beyond prediction and into planning**, but this requires knowledge bases of domain models and their cause and effect relationships, e.g. macro-economic models for global commerce and for recession. Where do we get them from? Later in the article we further discuss cause and effect reasoning.

(c) <u>**Thinking versus Calculating, Newton's Physics and DL**</u>. With data about billions of "things", and thousands of apples, that fall down every day, DL algorithms, with no human touch, can certainly come up with a model to calculate the time at which any given falling object will touch the ground. But can DL today come up with Newton's laws? I mean produce models and cause and effect statements that represent deeper generalized understanding of the Universe forces along with "compact" formula such as Time to hit the ground=SQRT(2*Height/g), where g=gravity=9.8 (as opposed to a gigantic black box neural net)? If you are only interested in building <u>calculators for smart machines</u>, you probably would ask back: "Why do I need a compact formula if I get the right result with a DL-generated 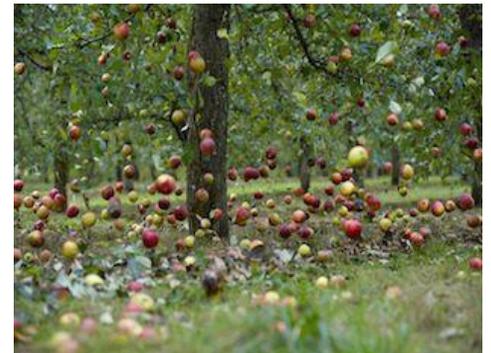 neural network (possibly even more accurate if the data also includes air resistance)?" My answer is that the **importance of deep understanding/grasping of Newton's laws (mid 1600's) goes way beyond a calculating tool for falling objects.** For example, the beautiful law of <u>inertia</u> is not at all about calculation: 'An object at rest (or in motion) remains at rest (or in motion), unless acted upon by an external force…'. Newton's laws describe processes of dynamic, continuous, infinitesimal interactions of bodies in the Universe. His discoveries were so deep, unique and innovative that he had to invent (or reinvent) the language of **calculus** because English, Latin or the mathematics that existed at his time could not express with clarity what he had in mind. **How are we going to represent such processes for AI?** Can we achieve that with Deep learning algorithms?

The abstraction, generalization, and analytic formula of Newton's laws were the knowledge base for future physicists to refine and expand them and discover new ones. Similar processes in the 1800's led to deeper understanding of heat and energy and the discovery of Thermodynamics laws, then Electricity, continuing all the way to Plank and Einstein (1900-1905) with Quantum Physics and relativity theory, and on to the engineering of the Apollo project and the recent missions to Mars and beyond. In most cases, science discoveries were (a) based on sparks of brilliant human theoretical thinking <u>with very little data</u> to learn from, as Sir *Isaac* Newton put it:" No great discovery was ever made without a bold guess", and (b) relied on what was already known; quoting *Newton* again*: "If I have seen further, it is by standing on the shoulders of giants".* As opposed to <u>learning from zero</u>, which is what a typical DL algorithm does.

**In summary, it is one thing to learn from data how to recognize visual objects or speech signals. It is a totally different challenge to derive from data full understanding of Newton's laws, how to build a bridge, or how equipment or human organs operate with their dynamic process flows.**

**(d) <u>Glaring Mistakes</u>**: Every other day we hear jokes about **glaring mistakes** by AI personal assistants, e.g. Siri, Cortana, Alexa, that quickly lead you to recognize the very limited understanding, scope and depth that the software has. DL developers typically focus on maximizing <u>overall</u> percentage accuracy, e.g. 94% correct classification. How do you protect a DL-based smart machine against glaring mistakes that even human beginners would not make? The more you push a DL algorithm for an <u>overall</u> higher percentage accuracy, the higher the likelihood that glaring mistakes will sneak in (one manifestation of overfitting). While in some applications glaring mistakes may be something to joke about, in others, e.g. military air-defense, they could be catastrophic. Just imagine **shooting down a passenger airplane mistakenly classified as a threatening object**. Similarly, for potential misclassification mistakes by autonomous land vehicles or drones. It only takes one or a few glaring mistakes to make users question the 'true' intelligence of the software to a point that the solution loses credibility and is soon rejected/shelved (before a costly or catastrophic event makes it too late). Arguing that the overall accuracy performance is within the, say 94%, promised error boundaries, does not help much.

Business software providers should specifically note that recovering from 'unforgivable' glaring-mistake events could take a long time and be very costly for the business. The client may activate liability clauses in the contract, and, worse, the competition may run a whole marketing campaign around it to destroy your product's reputation. That's why I always guide my teams with the following principle:
**Glaring Mistake Protection Principle**: In addition to working towards high overall accuracy, you should always include sanity checks to protect against glaring mistakes with individual cases. Before displaying the output of your 'ultra-intelligent' algorithm to a human user, or take automatic action, run sanity checks for extra protection.

Beyond preventing ordinary glaring mistakes, external knowledge could also contribute to overcoming mistakes due to intentional adversarial images designed to fool DL-based systems. The example in the next paragraph further illustrates the importance of external knowledge to complement data-only DL-based AI systems.

**(e) <u>External Knowledge which is not in DL's Data Set Could be Very Helpful:</u>** When a DL model does not produce good enough results, those who are radically religious about 'data-only' 'no human touch' doctrine adopt a closed garden end-to-end DL doctrine and limit their options to improve performance to those within the DL world, such as: add data, change neural net architecture, or augment the data with pre-processing operators, and then run again DL. Using external knowledge or alternative inference algorithms are taboo in this school. The following real-life example provides good reasons to reconsider the data-only doctrine:

**Example 2: Aerial Image Interpretation; a Twilight Zone Case.**
David and Abi- top notch experienced aerial image interpretation analysts at an Air-Force base- are faced with one of those challenging cases where they cannot decide whether an object O in an image is vehicle type A (agriculture) or B (military). The DL algorithm did not produce a conclusive recommendation, and they have tried all options to enhance the image for their own eyes, but uncertainty remains very high. Sam, a security specialist who happens to stop by just to say hello, arrives at the peak of their heated debate on A or B. As he is waiting, he looks at the picture and calmly says: "guys, coming from a farmer's family, and judging by the terrain and vegetation, I strongly doubt it is an agriculture vehicle, because no farmer would use A in this situation". After a short pause, Abi says "if this is B it might be a rocket launcher", and David replies: "Absolutely, I am going to wake up the boss".

Sam was using knowledge which is <u>not</u> in the picture data. Adopting the 'data only' DL doctrine is like David and Abi ignoring Sam's input. It limits the progress of DL. Well, a DL fan would now suggest collecting more data that cover Sam's farming knowledge and then re-run DL. That's a good theoretical exercise, but it does not go far, because vehicle B only shows up about twice a week for a short while, and every time with a slightly different silhouette, meaning about 100 pictures of B over a full year. Will this be sufficient for DL? Indeed, data augmentation can also be used to fight the low volume of data, and, yes, separate DL networks can be built to learn contexts where objects appear by terrain, vegetation, time of the year, etc., but **how about simply asking Sam?** I mean explicitly embedding farmers' knowledge in the AI solution. **Why wait to learn from data what humanity already know?**

(f) **When Sufficient Volume of Data does not Exist**. DL requires data, massive amounts of data, e.g. thousands of speech recording hours are needed to build a speech understanding system. For many business scenarios such volumes of data simply do not exist.

**Example 3: Troubleshooting a New Equipment**: Consider building an AI solution to support field service technicians of a new complex MRI (medical imaging) equipment that just came out. It may take several years before large and rich enough fault data become available for training a DL-based solution that can guide service technicians with efficient fault isolation and subsequent repair actions. **Do we not offer an AI solution until sufficient data is available for a DL-based solution?** In fact, by the time sufficient data is accumulated, the current equipment model is about to be replaced by a newer model. How would a DL algorithm know which data is only partially applicable, or no longer applicable? To further appreciate the low volumes of data relative to DL needs, keep in mind that, there are approximately 36,000 MRI machines worldwide from all vendors, on average, the total number of new MRI units sold annually from all vendors is roughly 2,500, they do not break every day, they have tens of different potential faults and hundreds of potential symptoms and test results, and they are typically replaced after seven years. http://www.magnetic-resonance.org/ch/21-01.html.

(g) **Explainable AI.** Today's DL solutions operate like a "black-box" and even their developers **cannot fully explain their reasoning**. For non-transactional applications in business, medicine, or military, explaining the reasoning is mandatory, or at least very highly desirable. DARPA' initiative into 'explainable AI' is very important [11].

**Where do we go from here?**

## 3. Double Deep Learning

The maximum a Deep Learning (DL) algorithm can learn is encapsulated in its data set for the given task, which, typically, is **substantially less than the full humanity's knowledge for the same task.** The **Double Deep Learning** approach advocates integrating data-centric **machine self-learning** techniques with **machine-teaching** techniques to leverage the power of both and overcome their corresponding limitations.
In the 'Double Deep' phrase, while the first 'deep' is for data-centric Deep Learning, the second 'deep' is for 'machine teachers' of knowledge with extra focus on teaching deep **'foundations'** and **'first principles'** for reasoning in the task domain. By 'deep' teaching I mean going beyond teaching shallow prescriptive knowledge about entities, e.g. WikiData, the CIA World FactBook, Google's and Microsoft's Knowledge Graphs, or just experiential rule-based knowledge with no model behind it. By analogy, I am talking about the difference between teaching **physicians** versus teaching **paramedics** or teaching **engineers** versus teaching **technicians.** Following Aristotle: "Knowledge of the fact differs from knowledge of the reason for the fact".

**Example 3 (continued) Equipment Troubleshooting:** Clearly, data about historic faults is a tremendous source for associating faults with symptoms and test results, but they can only take you that far in the equipment diagnostic process. Every field maintenance professional will tell you that being totally 'blind' about the equipment is the worst, and that **equipment-specific design knowledge**, e.g. its design diagrams, can greatly contribute to improve the troubleshooting process. Furthermore, to understand design diagrams, you use **universal engineering knowledge,** e.g. the generic function and behavior of fuses, RF amplifiers, and power supplies. The critical importance of knowledge beyond data is because **data can tell you how equipment fails, equipment-specific design knowledge and universal engineering knowledge can tell you how it works**. Both types of knowledge are required to reach high performance in diagnostic and repair decisions.
The AITEST system we deployed around the globe for dozens of complex large-scale equipment units [5], [6], illustrates automatic learning of massive scale Bayesian Inference Networks from engineering diagrams (structure and test paths) rather than only from data about historic faults of the equipment.

Table 1 summarizes some of the above key messages and illustrate that **knowledge plus data are likely to yield higher AI performance than data only**. See also Shoham [19] and Dietterich and Horvitz [10]. DARPA's important initiative into 'explainable AI' [11] is also likely to require adoption of the **Double Deep Learning** approach.

**Table 1: Data Tells you, ... Knowledge Tells you, ...**

| Object | Data tells you... | Knowledge tells you... |
|---|---|---|
| MRI (Medical equipment) | How it fails | How it works |
| Frigate ZX8 | What it looks like | Its capabilities |
| EKG signal | Arrhythmia type C | Relationship to heart's anatomy, physiology |
| Apples falling to the ground | How to calculate time to hit ground | Deeper understanding of Earth forces |

A key element of mankind's knowledge in just about every domain is cause and effect relationship. In the next section, we discuss cause and effect representation and reasoning as one area that further points at the synergy of data and knowledge and the potential of the Double Deep Learning approach.

## 4. Cause and Effect Reasoning and Learning

A cause and effect relationship typically emerges as a result of field observations and/or theoretic studies. Once conjectured, additional studies, e.g. carefully designed controlled experiments, are performed to prove or disprove it with a certain validity scope and likelihood. Some are deterministic, e.g. gravitation laws, some are probabilistic, e.g. high blood pressure is likely to cause heart failure. Some are easily derived by every human being, e.g. fingers in fire causes burns, some took many years, such as the decades it took to prove conclusively that smoking causes cancer. Cause and effect relationships may be quite complex via a network of multi-chain probabilistic connections where effect Y of cause X becomes the cause for effect Z. Not every cause and effect statement reaches consensus resolution like smoking, e.g. the current, yet undecided, debate on diet drinks and what aspartame may or may not cause. Also, causality relations and their explanation may be revised over time as mankind learns more, e.g. a new study shows that "exercise does not have to be prolonged in order to be beneficial. It just has to be frequent" (NY Times March 2018). Whatever the case, once a partial or full resolution of a cause and effect relationship is reached- which typically also involves peer reviews by independent experts- the conclusions, possibly not unequivocal- are added to the **state of the art of humanity knowledge bases** and **are used in real life decision making** such as court decisions in smoking-cancer cases, or providing bridges with safety certificates if certain potential risk-causes are eliminated.

The seminal works of Judea Pearl and others over the past 30 years, see [17] for extensive list of sources, offer the formalism needed to incorporate cause and effect relationships in smart machines. **But where do we take them from?** Option A is to take them from **mankind's knowledge repository of cause and effect relationships** in any specific domain and embed them explicitly in smart machines. Option B is to go back to humanity's darks days of ignorance and develop machine learning algorithms to re-discover from scratch every cause and effect relationship.

Learning cause and effect only from data with no human guidance is very challenging and should not be under-estimated. The well-known wisdom that "correlation does not necessarily imply causation" points to the complexity of the challenge to derive cause and effect by the current data-centric, self-learning, no-human-guidance algorithms. For example, the correlation between lightning and thunder was discovered thousands of years ago. Over time several theories were offered that lightning is the cause and thunder is the effect. However, the latest state of the art explanation how lightning causes thunder was only discovered in the 20th century by scientific reasoning leveraging modern physics knowledge, https://www.scientificamerican.com/article/what-causes-thunder/. Current machine self-learning-data-only algorithms could certainly predict the correlation between lightning and thunder from historic data, but lacking physics knowledge, how can they learn the true cause and effect relation? Similarly, in economics, machine learning algorithms can identify a large collection of correlations between recession-related variables, but without understanding the fundamentals of economic forces, we could not have reached the state of the art understanding of the causes of economic recession.

In a recent article, Pearl [16] eloquently points to inherent theoretical impediments of current data-centric machine learning algorithms for cause and effect relationships. I can certainly see the scientific merit of continued deep learning research to overcome these impediments, however, considering the benefits that AI systems can bring today, I would not wait until DL algorithms can learn the vast amount of all cause and effect relationships that mankind already knows. Note that even when a machine-learning algorithm converges to a cause and effect conclusion, validating it is far more demanding than the validation techniques currently used for classification and prediction by such algorithms. A faster way to scale the AI industry, is to embed the known cause effect relationships into what I call **Wikipedia for Smart Machines** (see below).

## 5. Wikipedia for Smart Machines Manifesto

**The Concept**

How about a repository that contains software representation structures of humanity's science & technology knowledge in various parts of life? Not just properties of things, correlations between things, cause and effect between things, but also deeper knowledge such as anatomy/physiology of the human organs, the way car engines work, what makes bridges safe, etc. Think about it like **Wikipedia for Smart Machines;** meaning target readers are not human, but rather smart machines. AI software developers will have these knowledge structures readily available, hence, the proposed name **ReKopedia,** where ReKo stands for reusable knowledge. I am not talking about a monolithic centrally managed initiative, but rather **a distributed self-organized initiative** managed by a mutually agreed upon governance. The technical representation structures in ReKopedia could be whatever we agree on: neural nets, Bayesian nets, cause and effect nets, mathematical formulas, sematic nets, knowledge graphs, state transition graphs, intelligent simulation models, logic, temporal logic, fuzzy logic, frames, or rules, as long as they are reusable by smart machines.

I am proposing a **community-wide initiative** to establish an **open-source shared knowledge repository** under which people contribute knowledge structures that are compatible with some protocols and enable others to use them. Past software initiatives in this direction, starting with the early days of Service Oriented Architecture (**SOA**), and on to **CORBA, REST** and beyond, can serve as a basis to learn from, e.g. self-contained modules, no need to know the inside technical details, and, of course, the concept of Web Services**.** Modularity is key, together with mechanisms to combine modules into higher level knowledge modules, which, when applied iteratively, create layers on top of layers of humanity knowledge.

In today's shared economy spirit, where software code for almost every AI algorithm can be found in open-source libraries like Python or R, the ReKo repository can be a significant complement and energize the AI industry. ***The same way we agreed to share open source software code, we will agree to share ReKo knowledge structures***. Note that knowledge sharing via textbooks and school teaching has already been a hallmark of mankind for generations. Let us do the same for smart machines.

**ReKopedia Content**

Content contribution to the **ReKopedia** repository can be made in any order and can come from everywhere, subject to some covenant rules. As people build smart machines for a variety of applications, they will contribute knowledge modules to the repository.

I can also envision special work-groups in different disciplines, e.g. Medicine, Economics, Agriculture, Environment, Military, Engineering, Manufacturing, Field Service, Finance, Insurance, Marketing, and Sales, each coming up with a long-term plan and priorities for the content that will populate the **ReKopedia** repository for their discipline. For example, compiling the state of the art knowledge about **inflation:** causes and effects, interventions and implications, and recovery patterns. Reviewing the **syllabuses of schools and universities** in different areas, will teach us the content we teach humans and can be a good starting point for the content we should teach machines. The Knowledge Graphs by Google and Microsoft including their API's illustrate such content, though they are currently focused on relatively shallow knowledge. The key principle, in my opinion, is to **maximize separation between the knowledge modules and the inference algorithms** that will operate on them, as Example 4 below illustrates for medical diagnosis.

**Example 4: ReKopedia Modules for Medical Diagnosis: Representation and Practice**
In medical schools we teach students (a) anatomy, physiology, and the like (b) characteristics of specific diseases/disorders by means of signs/symptoms/test results, along with sample cases, and (c) how to execute a good diagnostic process, generically. Note that (a) and (b) are knowledge modules, that are independent of (c) which is a generic inference process. In practice, when a patient arrives he/she does not announce "I have Appendicitis". He presents initial complaints/signs, and then a physician uses a generic inference engine (c), while accessing knowledge modules (a) and (b) to drive a cost-effective diagnostic process. That's the approach with which we built medical diagnostic systems for **endocrinology,** for **emergency and critical care (MEDAS)**, for **arthritis**, for **space medicine** and for **toxicology,** e.g. [2], [3]. The same inference engine was applied to knowledge modules of different medical fields whose knowledge were represented in the same way. (The anatomy and physiology part did not exist in the way I would do it today).

The Bayesian Network structures in Figure 1 illustrates the unified knowledge representation technique, while Figure 2 illustrates the set of generic inference algorithms we used for: **Hypothesis Generation, Goal Setting. Information Acquisition, and Evidential Reasoning**. Very similar knowledge representation and inference algorithms were used for several military situation assessment applications we developed [12], some have been in use for more than 20 years. The same concept of separating knowledge structures from inference algorithms was key to success.

### Medicine as a Case Study for ReKopedia Content

From years of experience, I learned that, on average, about **50 human expert hours** are needed to put into Bayesian Network templates the knowledge for diagnosing a single medical disorder. This means that with 50,000 hours we can complete 1,000 disorders which are likely to involve thousands of symptoms, signs, syndromes, test results, and other findings. Assuming 50 to 100 MD's working part time with support staff over 1 to 2 years, **a $15M budget would take care of the expenses to build ReKo modules for a non-marginal part of medicine.** In fact**,** the MEDAS approach advocates hierarchical structures of disorders, which means that after completing a base set, the average time per disorder will come down from 50 hours per disorder. MEDAS reached convincing performance in its early stages [4], and by 1990 it reached "**90% agreement with gold standard**" [12], long after I moved on to other areas. As for probability values on the links and nodes of the Bayesian Nets, we can start with known values from medical publications and data bases, or expert's subjective values, and then, as more patient data is accumulated, apply machine learning algorithms to update them. In [7] we report on sensitivity analysis research that shows that Bayesian Net evidential reasoning tolerate fairly large deviations in the prior and conditional probabilities without impacting correct diagnoses. (Remember that AI is not about pure probability calculations. It is about converging to the right diagnosis. If a patient suffers from disorder D, then an intelligent diagnostic process will lead you to collect the right findings that eventually will bring D's probability to the top). The Bayesian Nets should be complemented and connected with ReKo modules that capture knowledge about **anatomy, physiology, bio-engineering, DNA**, and **other knowledge sources** that can improve the inferences made by the Bayesian inference algorithms. Looking not too far in the future, when in large parts of the world patient data is automated starting at birth date, including genome map data for every individual, ReKopedia-based AI systems can take intelligent healthcare automation to new heights in terms of early warning, prevention, diagnosis and treatment, reducing cost while improving quality.

### Automating the Generation of ReKopedia Content

One of the key goals of the ReKopedia initiative is to develop automatic and semi-automatic algorithms to convert humanity's documented knowledge into software structures that smart machines can use in their inference algorithms. Automatic conversion of natural language material; including diagrams and pictures, into ReKo structures, can accelerate considerably ReKopedia development, but it requires taking Natural Language Understanding (NLU) to a higher level. We are not there yet. To get an appreciation of the challenge, consider the task of automatic summary generation of text documents. The state of the art of this field tells us that AI NLU software is still far away from "truly understanding" what it reads, let alone extracting knowledge from it, as compared with a college student reading a chapter in an Economics 101 textbook and being able to solve homework exercises. This should not stop us, however, from starting to manually build the **ReKo** repository for any field of science and technology we desire.

Figure 1: MEDAS- Hierarchical Structure of Medical Disorders [2]

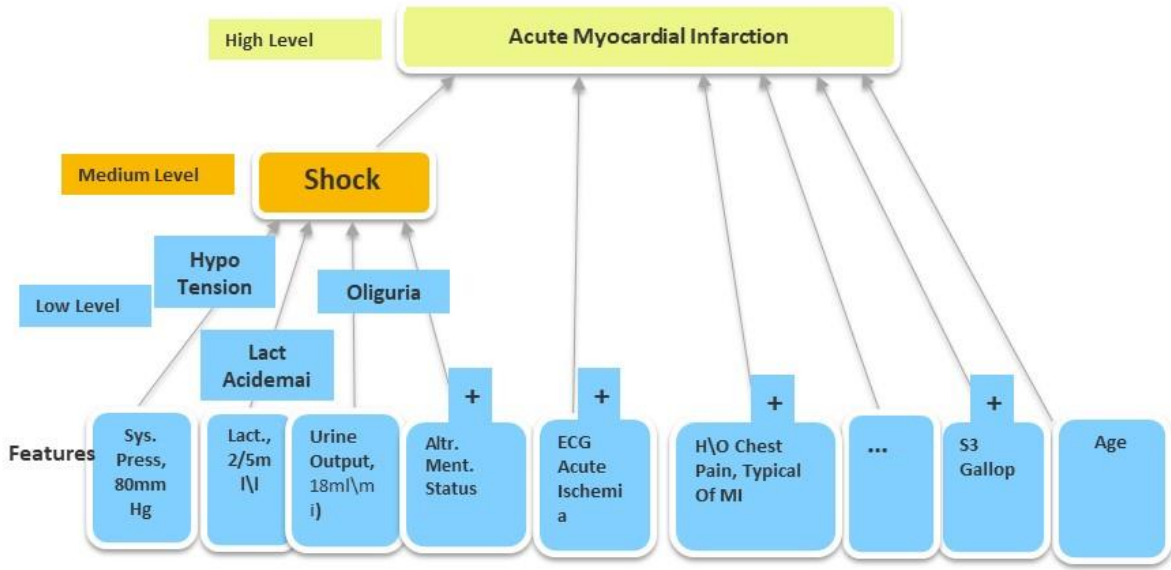

Figure 2: Cycle of Diagnostic Assessment, BenBassat 1980's [2], [3]

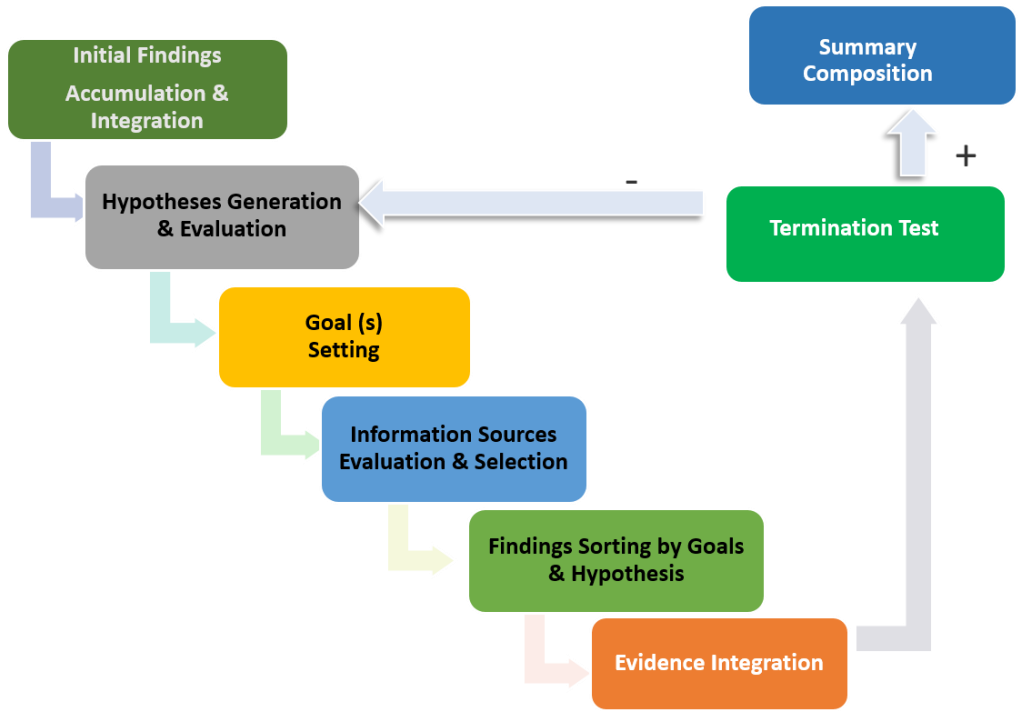

## 6. Summary

Today's DL-based AI applications are typically point solutions for transactional tasks that do not lend themselves to automatic generalization beyond the scope of the data sets they are based on. Non-transactional applications require AI algorithms that integrate deeper and broader knowledge in their problem-solving capabilities. The proposed **Wikipedia for smart machines (ReKopedia)** along with the **double deep learning** approach offer the paradigm for integrating data-centric 'knowledge blind' algorithms with deep-knowledge-driven algorithms. The resulting synergies will establish broad and deep foundations that will enable us to achieve higher level 'generic', 'universal' machine intelligence and scale faster the AI industry.

I hope that people will approach the ReKopedia idea in a pragmatic manner rather than with the classic philosophical discussions of 'what machines can/cannot do'. There is so much we can surely achieve today with the ReKopedia approach beyond what can be achieved with the data-centric 'knowledge blind' approach.

**We are now in the second AI 'spring' after a long 'winter'. To avoid sliding again into an AI winter, notwithstanding the question whether deep learning is or is not alchemy, it is essential that we rebalance the roles of data and knowledge. Data is important but knowledge at all levels: shallow, deep and commonsense are equally important.**

**If indeed AI is the driver of our next economic and social revolution (like electricity was), we'd better establish solid foundations and infrastructure to develop and disseminate it with standards and fair economics.**

## Appendix: Key Messages

1. Machines can learn many things from data, but data is not the only source machines can learn from.

2. The maximum knowledge a Deep Learning (DL) algorithm can learn is what is encapsulated in its data set for the given task, which, typically, is substantially less than the full humanity's knowledge for the same task. For instance, using field service data, a DL algorithm can learn the symptoms an equipment shows when it fails, even predict when it is likely to fail, but DL cannot learn from such data how it works.

3. Developing and deploying AI solutions when data is not (yet) available is possible, with substantial business value, by directly embedding explicit humanity knowledge.

4. It is one thing to train a DL-neural-net to calculate the time a falling apple will hit the ground using data about falling apples. It is a totally different challenge to train an algorithm from data only to come up with Newton's gravity laws. **If humans can learn from explicit teaching, why can't machines?**

5. Current data-centric machine learning technologies have inherent limitations in learning certain problem-solving structures, e.g. Cause and effect. Why wait until- and if- an algorithm is developed to learn from data what humanity already learned and documented in textbooks and other publications? E.g. anatomy, physiology, engineering, etc. **Science-wise it has merit. Business-wise it makes no sense**.

6. In today's AI world, **data is over-rated, knowledge is under-rated.** By re-balancing the two, AI solutions will benefit considerably. On the other hand, by adopting a 'data-only' doctrine, you give up many options to improve AI performance and expand its applicability.

7. The **Double Deep Learning** approach advocates integrating 'machine self-learning' with 'machine teachers' with extra focus on teaching deep 'foundational' 'first principles' knowledge aiming at higher level intelligence, like the difference between teaching **physicians** versus **paramedics**, or teaching **engineers** versus **technicians**.

8. **Wikipedia for smart machines**. AI can grow faster by establishing an **open-source shared repository of reusable knowledge modules** (coined **ReKopedia** here) covering humanity's science & technology in various disciplines. For illustration, a $15M project is proposed to produce ReKopedia modules for medical diagnosis of 1,000 disorders.

## *Personal Note, Acknowledgments & Short Bio

This article is based on decades of my AI-focused career that are a blend of being a mathematician/statistician/computer scientist (USC, Tel-Aviv University, UCLA), and being a business entrepreneur and 15 years CEO of ClickSoftware, a NASDAQ (CKSW) public company. I have been researching, practicing and educating Artificial Intelligence from the first AI "Spring" of the 1980's, during the AI "Winter" of the 1990th and early 2000 years, and now in the AI renaissance of the 21th century. My academic research was supported by NIH, NSF, DARPA, NASA, BMD (Ballistic Missile Defense Agency), ARI (U.S. Army Research Institute), Israel Defense Forces, and others.

As the founder and CEO of ClickSoftware (inventor of service chain optimization- patent awarded, acquired in 2015 by a private equity firm), and Plataine (a leader in **AI and IoT**-based solutions for manufacturing optimization), we leverage AI technologies to solve **large scale real-life business problems**. We developed innovative AI products that benefit hundreds of millions of people around the globe.

For more details and Publication List see **http://www.moshebenbassat.com/.**

**I am very grateful to Israel Beniaminy, my son Avner Ben-Bassat, and other colleagues for deep and useful discussions, as well as comments on early drafts of this article. Also thanks to Judea Pearl for calling my attention to important developments in causality.**